\documentclass[runningheads]{llncs}
\usepackage{graphicx}
\usepackage[numbers, sort&compress]{natbib}

\usepackage{xcolor}
\usepackage{floatrow}
\usepackage{fancyhdr}
\usepackage{comment}

\fancypagestyle{FirstPage}{
\lfoot{\textit{Peer-reviewed short paper presented at 5th International Workshop on Cognition:Interdisciplinary Foundations, Models and Applications,(CIFMA'23).}} 
}

\begin{document}
%
\title{Toward enriched Cognitive Learning with XAI}

%
\titlerunning{Toward enriched Cognitive Learning with XAI}

%
\author{Muhammad Suffian\inst{1}
\orcidID{0000-0002-1946-285X}
Ulrike Kuhl\inst{2}
\orcidID{0000-0002-9405-918X}
Jose M. Alonso-Moral\inst{3}\orcidID{0000-0003-3673-421X}
\and
Alessandro Bogliolo\inst{1}\orcidID{0000-0001-6666-3315}
}
\authorrunning{M. Suffian et al.}
%
\institute{Department of Pure and Applied Sciences, University of Urbino, Urbino, Italy\\
\email{m.suffian@campus.uniurb.it, alessandro.bogliolo@uniurb.it}
\and
Research Institute for Cognition and Robotics,
Bielefeld University, Bielefeld, Germany\\
\email{ukuhl@techfak.uni-bielefeld.de}
\and 
Centro Singular de Investigaci\'{o}n en Tecnolox\'{i}as Intelixentes (CiTIUS), Universidade de Santiago de Compostela, 15782 Santiago de Compostela, Spain\\
\email{josemaria.alonso.moral@usc.es}
}
\maketitle            

\begin{abstract}
As computational systems supported by artificial intelligence (AI) techniques continue to play an increasingly pivotal role in making high-stakes recommendations and decisions across various domains, the demand for explainable AI (XAI) has grown significantly, extending its impact into cognitive learning research. Providing explanations for novel concepts is recognised as a fundamental aid in the learning process, particularly when addressing challenges stemming from knowledge deficiencies and skill application. Addressing these difficulties involves timely explanations and guidance throughout the learning process, prompting the interest of AI experts in developing explainer models. In this paper, we introduce an intelligent system (CL-XAI) for Cognitive Learning which is supported by XAI, focusing on two key research objectives: exploring how human learners comprehend the internal mechanisms of AI models using XAI tools and evaluating the 
effectiveness of such tools through human feedback. The use of CL-XAI is illustrated with a game-inspired virtual use case where learners tackle combinatorial problems to enhance problem-solving skills and deepen their understanding of complex concepts, highlighting the potential for transformative advances in cognitive learning and co-learning.
\end{abstract}

\keywords{Cognitive Learning \and Explainable AI, \and Human-centered AI \and Problem Solving \and Counterfactual Explanations \and Co-Learning.}

\section{Introduction}
\thispagestyle{FirstPage}

In the realms of learning theory, artificial intelligence (AI), and human-computer interaction (HCI), the pursuit of problem-solving and optimal solution finding 
have historically been perceived through distinct lenses for machines and humans \cite{langley2006intelligent, villaronga2018humans, zhang2008cognitive}. Machines, equipped with their computational capabilities and cost-driven optimisation, operate in a space detached from human cognition \cite{newell1972human}. Conversely, humans rely on their unique problem-solving approaches, drawing from experiences and intuition \cite{newell2007computer}. This separation contradicts the principles of co-learning and effective 
HCI. In our contemporary era, where AI systems often outperform humans in various domains, bridging this gap is imperative to create a more enriching learning environment \cite{grace2018will}.
The underexplored domain of human-machine co-learning, aiming to foster mutual improvement and progress, must be addressed. Humans need an opportunity to plunge into the intricate inner workings of AI's cognitive machinery, actively participating in co-learning to solve complex problems. In turn, AI systems should reap the benefits of human wisdom, leveraging human input and feedback to alleviate the computational burdens that once restrained their problem-solving prowess.

As 
AI systems 
play an increasingly pivotal role in making high-stakes recommendations and decisions across various domains, the demand for eXplainable 
AI (XAI) to elucidate the rationale behind these systems 
grows 
\cite{adadi2018peeking}. 
This paper introduces 
the so-called Cognitive Learning 
with eXplainable 
AI (CL-XAI) system.
We explore the potential of co-learning with counterfactual explanations (CEs),
where humans and machines collaborate in problem-solving tasks. CEs
enable users to grasp the ``what if" aspect of AI decisions, shedding light on alternative courses of action and improving 
transparent communication \cite{wachter2017counterfactual}.

In this context, our focus is twofold: firstly, we investigate how human learners' cognitive learning processes are influenced when they use the CL-XAI tool to receive explanations. Secondly, we rigorously assess the effectiveness of the CL-XAI tool by gathering feedback from humans to evaluate its accuracy and helpfulness in providing explanations. The co-learning experience unfolds with regular interactions between learners and an explanation tool. This tool bridges human intuition and machine logic, offering learners a lifeline in their pursuit of optimal solutions. 
If a human learner successfully identifies 
an optimal solution, then it is an evidence
to the acquisition of knowledge comparable to that acquired by a machine learning (ML) model \cite{ribeiro2016model}.
For the co-learning experience, we propose a virtual game-inspired scenario where learners solve combinatorial problems to achieve improved results (see section~\ref{clxai}). The learner 
receives explanations at regular intervals to solve the task, and the log of the learner's choices and attempts is recorded to evaluate 
its mental model. Overall, we believe that the synergy between human cognition and XAI
guidance holds the promise of transformative advances in cognitive learning, a step towards co-learning, ultimately bolstering problem-solving skills and fostering a comprehensive understanding of complex concepts. 

The rest of the paper is structured as follows. Section~\ref{background} provides the necessary background. Section~\ref{clxai} outlines the CL-XAI 
system, elucidating our approach to incorporating XAI into the cognitive learning process.
Section~\ref{subjectiveEValuations} delves into subjective evaluation measures for the proposed framework, and finally, in section~\ref{conclusion}, we draw conclusions from our work, emphasising 
potential applications and the avenues for future research in the realm of XAI-driven cognitive learning.

\section{Background}\label{background}

Cognitive learning is a pedagogical approach that emphasises the development of comprehensive mental models among learners \cite{johnson2013mental}. Such mental models play a pivotal role in knowledge acquisition and problem-solving, enabling individuals to navigate complex domains effectively \cite{vanlehn1996cognitive}. The collaboration between humans and machines in cognitive learning, often called co-learning, has
great potential 
\cite{deiss2018hamlet, lieto2016human}. 
It has been shown to enhance problem-solving skills and deepen the understanding of complex concepts \cite{lieto2016human}.
On the one hand, prior studies have introduced theoretical frameworks for symbiotic learning systems \cite{wu2021analytics}. These frameworks depict a reciprocal learning process, where the learner acquires knowledge from the system, and, conversely, the system gains insights from the learner, facilitated through reinforcement learning.
On the other hand, research into explainable recommendation systems has expanded to education. For instance, Barria-Pineda et al. \cite{barria2021explainable} have explored the domain of recommending resources in programming classes. Tsiakas et al. \cite{tsiakas2020brainhood} have investigated using cognitive training recommendations for primary and secondary school children.

In addition, in the field of mathematics \cite{renkl2017learning} and across diverse scientific domains, including chemistry \cite{crippen2007impact}, educators routinely employ worked examples. These pedagogical tools serve a dual purpose: they provide solutions and offer explanations rooted in an expert's mental framework, which can be comprehended from a novice's standpoint \cite{sweller2006worked, suffian2022human}.
Conversely, our approach integrates an XAI tool for generating explanations, which are subsequently provided to novice learners to facilitate the construction of their mental models. This method constitutes an automated system that circumvents the necessity for an expert's mental model, as it leverages the pre-existing capabilities of the XAI system.

\section{CL-XAI}\label{clxai}

CL-XAI is a 
tool that encompasses three different components: 
(1) 
The explanation method called User Feedback-based Counterfactual Explanations (UFCE) \cite{suffian2022fce} to assist 
learners in solving a given combinatorial problem. 
(2) The 
virtual \textit{Alien Zoo} framework \cite{kuhl2023AlienZoo} is utilised to design the task for the learners as a use case. 
(3) A web-based game-inspired user study is designed for 
learners
to enhance their learning and knowledge about the artefacts of the underlying AI model when solving the given combinatorial problem.  

\subsection{UFCE}
The field of XAI encompasses various technical approaches aimed at enhancing the transparency, interpretability and explainability of AI systems~\cite{jose2023wwwxait}. One prominent method involves using local post-hoc interpretability approaches for elaborating explanations of black-box models.
Such approaches are usually supported by inherent interpretable white-box models such as 
linear models, decision trees or rule-based systems.
Indeed, providing users with explicit rule sets governing AI behaviour is a way of ensuring a clearer understanding of the decision-making process. Additionally, techniques such as feature importance scoring, attention mechanisms, and saliency maps enable the identification of crucial model elements and their contributions to the overall outcome. CEs provide insights into how alterations in input data affect the model's outputs, while visual aids like heatmaps and activation maps highlight influential areas in images. Moreover, human-in-the-loop interactions and natural language explanations also contribute to the holistic understanding of AI operations, catering to various interpretation needs and promoting trust in AI decision-making.

In this context, UFCE\footnote{https://github.com/msnizami/UFCE} provides CEs\footnote{In the rest of the paper, we use explanations and counterfactual explanations (CEs) interchangeably.} to the learner to devise different strategies to solve the task at hand. UFCE, herein referred to as the ``XAI tool'', is not self-driven; rather, it is a human-in-the-loop explainer that exploits learner feedback to generate customised explanations. In addition, it guides the learner towards correct input options whenever the given feedback does not produce any explanation. 
The task is designed by keeping the needs of both the learner and the explainer.

\subsection{Use Case and Experimental Context}

The efficacy of an explanation is 
influenced by its purpose and the specific audience it is intended for \cite{mohseni2021multidisciplinary}. These two factors are pivotal in selecting a suitable use case and shaping the experimental environment. Thus, we have adopted the Alien Zoo \cite{kuhl2023AlienZoo}, a web-based game-inspired virtual space wherein the learners do not have prior knowledge and only gain knowledge through the explanations provided by the explainer.

The learner's task is to nurture an Alien so-called `Shub' in the Alien Zoo by feeding various combinations of plants. In our case, in contrast to the original implementation \cite{kuhl2023AlienZoo}, the well-being of Shub is directly influenced by the choices learners make in selecting plant combinations. The leaves of plants are attached with a specific time cost to find them. The learner is exposed to a combinatorial problem in which the random number of leaves of each plant are given, which are not a better diet for the health of Shub. The learner has to solve this problem by making different combinations of the plants to constitute a nourishing diet adhering to their time cost. The learner-selected plants will be fed to 
an AI system, which will predict whether this combination can improve Shub's health. 
The choices made when feeding Shub have immediate outcomes, resulting in poor or better health. The XAI tool guides the learner in making optimal decisions. At regular intervals, learners are provided with explanations alongside their previous selections. These explanations highlight a choice that could have yielded a more favourable outcome. Also, these explanations are, by default, enriched with hints about the underlying data distribution 
associated to the prediction model and about how Shub's health improves, ultimately assisting 
learners in making informed decisions and 
enhancing their mental model.

\subsection{Game-inspired User Study}

\begin{figure}[hbt!]
    \centering
    \includegraphics[width=.7\textwidth]{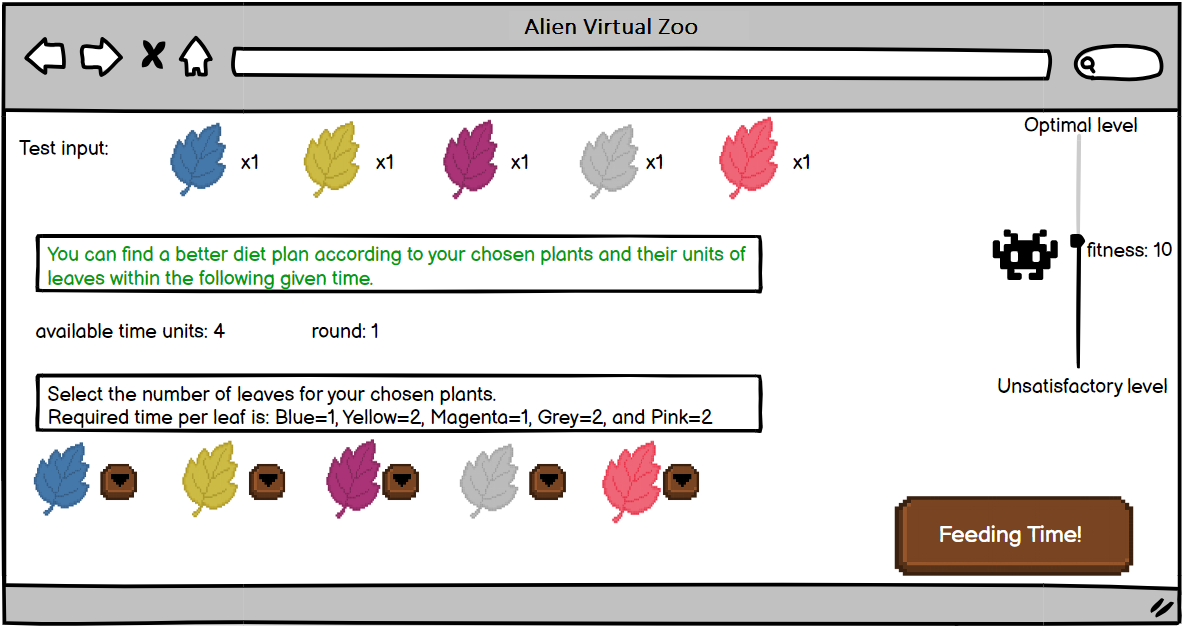}
    \caption{The learner's task is to go for those combinations of plants by selecting from drop-down menus that can be searched in the available time. After each search/selection, the learner feeds the plants to Shub and waits for the outcome (which is displayed in the next screen, see Fig.~\ref{fig:screen-1A-3}).}
    \label{fig:screen-1A}
\end{figure}

A game-inspired user study is designed to record the user solutions for the given tasks and explanations provided at regular intervals. The data is collected through game locks (logs) for the analysis and evaluation of the cognitive learning process of 
learners and the extent of goodness of provided explanations by the XAI tool.
The 
user interface (UI) is meticulously designed to offer learners an intuitive and engaging experience throughout their journey of nurturing Shub and exploring the intricate relationships between plant combinations and growth outcomes. The UI comprises various screens, each carefully crafted to provide essential information and interactivity. In the right hand side of Fig.~\ref{fig:screen-1A}, an avatar represents Shub and dynamically reflects its fitness level on a vertical bar as the chosen plant combination impacts Shub's health. The bar limits indicate the optimal and unsatisfactory fitness levels. On the left, the five different plants are displayed on top, which constitute the diet, and a test input is shown at the bottom, which needs the learner's attention to customise it by selecting any combination from the drop-down menu's given for each plant. 
The user can also see the available time to improve the health with the number of rounds and the time required to invest in one leaf of each plant. 

\begin{figure}[hbt!]
    \centering
    \includegraphics[width=.7\textwidth]{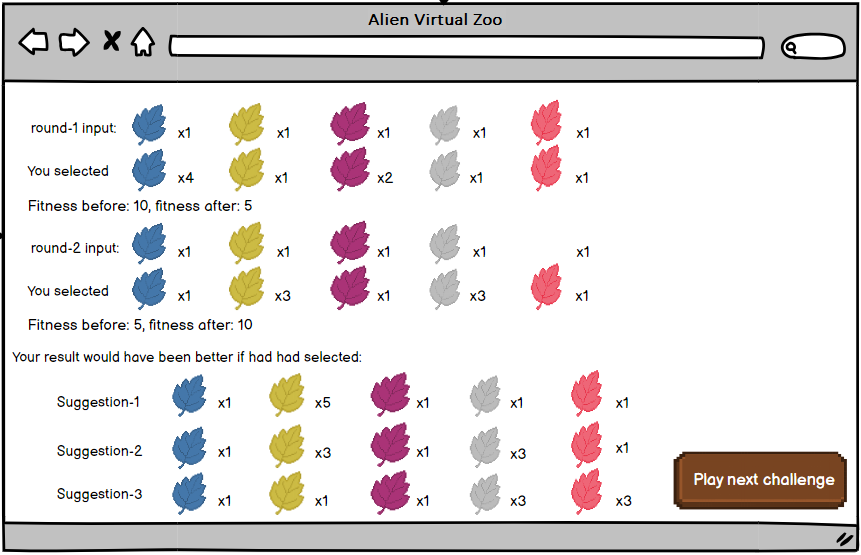}
    \caption{In this screen, the learner 
    can see the selection of plants throughout the different rounds and their outcome for fitness (better or worse). Additionally, the learner 
    receives different suggestions (explanations). If the learner had used these suggestions, this would have been a better diet for Shub.}
    \label{fig:screen-1A-3}
\end{figure} 

\textbf{Technical Details.}
The realisation of the Alien Zoo entails a rigorous segregation between the front end, responsible for crafting the game interface that participants interact with, and the back end, delivering the 
predictions made by the AI system
along with the explanations provided by the XAI tool.
The web interface utilises Phaser3\footnote{https://phaser.io/}, an HTML5 game framework driven by JavaScript. The system's backend, on the other hand, is founded on Python3, leveraging the sklearn\footnote{https://scikit-learn.org/stable/} package for supporting ML 
algorithms.
The underlying ML model is 
trained using synthetic plant data \cite{kuhl2023AlienZoo},
and it predicts the fitness of Shub, thereby influencing the game's dynamics. The learner input reaches this model via the front end, allowing an analysis of the potential for yielding positive outcomes and consequently enhancing Shub's fitness.
This Python-based framework is adept at generating CEs with UFCE to ensure adaptability and 
accommodate various 
ML algorithms.

\section{Subjective Evaluation Measures}\label{subjectiveEValuations}

In evaluating XAI systems, examining specific cognitive states or processes, herein referred to as ``cognitive metrics", 
is a central concern \cite{cogmetrics}. These metrics are instrumental in assessing whether learners have achieved a pragmatic understanding of the AI system, particularly in light of the explanations furnished by the XAI tool. Drawing from established approaches in cognitive science and psychology \cite{cogmetrics}, Hoffman et al. \cite{hoffman2018metrics} proposed a conceptual model elucidating the explanation processes of an XAI system and how a learner's pragmatic comprehension of these explanations can be assessed across distinct functional stages. Hoffman's model delineates three pivotal functional stages within the XAI system's operations: \textit{explanation generation}, \textit{learner's mental model generation}, and \textit{learner's enhanced performance resulting from the assimilated mental model}. Consequently, the evaluation of XAI systems through cognitive metrics can be systematically framed within these three stages, offering a comprehensive approach to gauge the effectiveness of these systems in facilitating user understanding and performance improvement.

In accordance with Hoffman et al. \cite{hoffman2018metrics}, during the explanation generation phase, an assessment of the learner's practical comprehension of the AI system can be made by examining the learner's cognitive processes, which gauge the quality of the explanation (referred to as ``Explanation Goodness") and the degree of satisfaction with it (referred to as ``User Satisfaction"). Between the explanation generation phase and the subsequent stage of constructing the learner's mental model, which is influenced by the explanations received, learners gradually 
update their mental models, with several psychological factors potentially influencing 
the model-building process. Evaluating the extent of a learner's understanding and satisfaction poses a formidable challenge. To address this challenge, we have 
extended the Alien Zoo, where the learners are now assigned problem-solving tasks, and their performance in solving these tasks serves as a metric to determine the level of understanding of the provided explanations. This, in turn, corroborates the effectiveness of the XAI tool in generating high-quality explanations.
User Satisfaction can be assessed using subjective measures through the administration of a questionnaire, with for example the \textit{Explanation Satisfaction Scale} introduced by Hoffman et al. \cite{hoffman2018metrics} and later refined by van der Waa et al. \cite{vanderwaa2021evaluating}. This scale provides a reliable and psychometrically robust means of gauging user satisfaction with a system's explanations.

User understanding, in the context of XAI, pertains to the development of a learner's ``mental model" of a system's inner workings \cite{cogmetrics}. The concept of a ``mental model" draws from psychological theories, denoting an individual's internal representation of the people, objects, and environments with which they interact \cite{staggers1993mental, richardson1994foundations}. In the realm of XAI, for our case, an ideal outcome is for the learner's mental model to reflect the XAI tool accurately. Explanations play a pivotal role in facilitating the construction of precise mental models, which can be categorized as follows: global understanding, signifying a general comprehension of a system's functioning; local understanding, signifying insight into a specific decision made by the system; and functional understanding, representing a grasp of the system's capabilities and intended uses \cite{hoffman2018metrics, cogmetrics}. 
In this work, our goal is to capture the local understanding of the learner's mental model.

In cognitive psychology research, learners' performance in task-learning endeavours is subject to the influence of many variables. These encompass factors such as age \cite{chan2018eye}, learning experience \cite{villaronga2018humans}, cognitive abilities \cite{johnson2013mental}, cultural and socio-emotional factors \cite{varnum2010origin}.
Factors related to mental health \cite{marin2011chronic} collectively constitute critical determinants. 

\textbf{Proof of Concept}
In our 
ongoing user study, 
the focus is on cognitive learning and co-learning mechanisms. We aim to investigate 
4 key measures: 
\begin{itemize}
 \item \textbf{Explanation Goodness}. We evaluate the quality of explanations provided to learners, aiming to assess how well they convey complex data relationships. We anticipate that well-crafted explanations will positively influence other measures.
 
 \item \textbf{User Satisfaction}. Post-game surveys collect learner feedback on their satisfaction with the explanations. Learners' assessments of the explanations' comprehensibility, usefulness, and overall satisfaction provide insights into user satisfaction.
 
 \item \textbf{User Understanding}. Our primary inquiry is whether users can enhance their understanding of complex data relationships through explanations. By providing CEs, we aim to help users better grasp the system's intricacies.
 
 \item \textbf{Task Learning}. We expect that improved user understanding, facilitated by the explanations, will lead to task learning. Task learning is assessed through metrics such as fitness levels during the game and the time taken to make decisions, reflecting learners' increased ability to identify crucial data factors and make informed choices.
\end{itemize}
By examining these measures, we aim to uncover how explanation quality influences cognitive learning and co-learning mechanisms 
with the CL-XAI 
tool.

\section{Conclusion}\label{conclusion}
In this paper, we have introduced 
the CL-XAI 
tool, designed to facilitate cognitive learning
with XAI. The outcomes of our research will offer significant potential for enhancing the development and refinement of XAI 
techniques, ultimately leading to improved cognitive learning experiences.

Our deliberate focus on the convergence of XAI and cognitive learning stems from the recognition that learners, especially in educational and training contexts, stand to gain substantially from AI systems that are understandable and transparent. Our work, centred on providing timely and lucid explanations for intricate concepts and problem-solving tasks (combinatorial problems), seeks to empower learners, bridge knowledge disparities, and cultivate a deeper understanding of challenging subject matter.
Furthermore, the implications of 
this research extend beyond the realm of education to encompass domains where human-AI collaboration is pivotal, such as healthcare diagnostics, legal decision support, and financial analysis.

In summary, we assert that the synergy between human cognition and the guidance offered by the CL-XAI tool holds the potential for transformative advancements in cognitive learning. This will mark a significant stride toward co-learning, ultimately fortifying problem-solving abilities and nurturing a comprehensive grasp of complex concepts.

\section*{Acknowledgments}
This research was funded by MCIN/AEI/10.13039/501100011033 (grants PID2021-123152OB-C21 and TED2021-130295B-C33), the Galician Ministry of Culture, Education, Professional Training, and University (grants ED431C2022/19 and ED431G2019/04). All grants were co-funded by the European Regional Development Fund (ERDF/FEDER program). Ulrike Kuhl was supported by the research training group ``Dataninja'' (Trustworthy AI for Seamless Problem Solving: Next Generation Intelligence Joins Robust Data Analysis) funded by the German federal state of North Rhine-Westphalia.
%
%
%
\bibliographystyle{splncs04}
{\footnotesize
\bibliography{ref}}
%

\end{document}